\documentclass{article}
\usepackage{interspeech_04,amssymb,amsmath,epsfig}
\def\reg{{\rm\ooalign{\hfil
     \raise.07ex\hbox{\scriptsize R}\hfil\crcr\mathhexbox20D}}}

\title{Effects of Language Modeling on Speech-driven Question Answering}

\makeatletter
\def\name#1{\gdef\@name{#1\\}}
\makeatother
\name{{\em Tomoyosi Akiba $^\dag$, Atsushi Fujii$^\ddag$, Katunobu Itou$^*$}}

\address{$^\dag$ 
  Toyohashi University of Technology\\
  1-1 Hibarigaoka, Tenpaku-cho, Toyohashi, 441-8580, JAPAN\\
  {\small \tt akiba@cl.ics.tut.ac.jp}\\ %\\
  $^\ddag$ 
  University of Tsukuba\\
  1-2 Kasuga, Tsukuba, 305-8550, JAPAN\\
  $^*$ 
  Nagoya University\\
  1 Furo-cho, Nagoya, 464-8603, JAPAN\\  %\\
}

\begin{document}
\maketitle

\begin{abstract}
  We integrate automatic speech recognition (ASR) and question
  answering (QA) to realize a speech-driven QA system, and evaluate
  its performance.  We adapt an N-gram language model to natural
  language questions, so that the input of our system can be
  recognized with a high accuracy. We target WH-questions which
  consist of the topic part and fixed phrase used to ask about
  something. We first produce a general N-gram model intended to
  recognize the topic and emphasize the counts of the N-grams that
  correspond to the fixed phrases. Given a transcription by the ASR
  engine, the QA engine extracts the answer candidates from target
  documents. We propose a passage retrieval method robust against
  recognition errors in the transcription. We use the QA test
  collection produced in NTCIR, which is a TREC-style evaluation
  workshop, and show the effectiveness of our method by means of
  experiments.
\end{abstract}

\section{Introduction}

Question Answering (QA) was first evaluated extensively at TREC-8
\cite{voorhees/1999/trec8}. The goal in the QA task is to extract
words or phrases as the answer to a question, rather than the document
lists obtained by traditional information retrieval (IR) systems.
Speech interfaces have promise for improving the utility of QA
systems, in which natural language questions are used as inputs. We
enhanced our speech-driven IR system \cite{fujii:springer-2002} to
accept spoken questions.

In this paper, we evaluate the effects of language modeling on
speech-driven question answering. In past literature, language models
were evaluated independent of specific tasks.  Perplexity is one of
the common measures to evaluate language models, irrespective of the
speech recognition accuracy.
Word error rate (WER) is another common measure, which directly
evaluates the accuracy of speech recognition.  However, it is not
clear that they can evaluate the performance of specific
information processing systems using speech interfaces.  Because
question answering is one of the well-defined tasks and has been
evaluated by formal evaluation workshops, e.g., TREC and NTCIR, we can
evaluate components of a system, in particular language modeling,
through a rigorous method.

Section \ref{ss:language_model} describes our language modeling method
for speech-driven question
answering \cite{akiba/2003/adapting}. Section
\ref{ss:question_answering} describes our question answering
engine \cite{akiba/2004/question}. Section \ref{ss:evaluation}
describes the experimental results.

\section{Language Modeling for Question Answering}
\label{ss:language_model}

Question answering systems accept a question consisting of the part
that conveys a topic and the part that represents a fixed phrase for
question sentences. The following is an example question:
\begin{quote}
  {\it seN / kyu-/ hyaku/ nana / ju- / roku / neN / ni / kasei / ni / naN / chakuriku / shita / taNsaki / wa / naN / to / yu- / namae /desu / ka}\\
  (What was the name of the spacecraft that landed safely on Mars in
  1976 ?)
\end{quote}
The first half of the question, i.e., ``{\it seN kyu- hyaku nana ju-
  roku neN ni kasei ni naN chakuriku shita taNsaki wa} (the spacecraft
that landed safely on Mars in 1976)'', conveys the topic, and can be
recognized by an N-gram model trained with target documents (e.g.,
newspaper articles). The latter half of the question, i.e., ``{\it naN
  to yu- namae desu ka} (What was the name?)'', is a fixed phrase
typically used in interrogative questions, which is not very frequent
in newspaper articles.  Thus, we need a language model adapted to both
types of expressions.

Note that recognizing the fixed phrases with high accuracy is crucial
in question answering, because these phrases convey clues to determine
the question and answer types. For example, a fixed phrase indicates
that the answer should be the name of an object as in the previous
example, while another question can potentially indicate that the
answer should be the date of an event (e.g., ``On what date was...'').

In this paper, we use our previous method \cite{akiba/2003/adapting},
in which a language model for question answering are produced from a
list of the fixed phrases typically used in interrogative questions.
This method emphasizes the N-gram subset corresponding to the fixed
phrases. This method can be recast as a variant of maximum a posteriori
probability (MAP) estimation, in which the N-gram subset of a
background corpus is used as a posterior distribution.

\subsection{Language Modeling by Emphasizing N-gram Subsets}

Let $S$ be a set of sentences.  Let $S_{FP}$ be a subset of $S$ that
consists of the sentences including the fixed phrases in a list. Let
$P$ be a language model of generating sentences (i.e., $s \in S$)
obtained from a general-purpose background corpus. The aim of the
language model adaptation for the fixed phrases is to obtain the
adapted language model $P'$, which provides higher probability scores
for sentence $\hat{s} \in S_{FP}$ but maintains the order relations on
sentences $s \in S - S_{FP}$ as much as possible.

The adapted model $P'$ is produced by the following two steps.
\begin{enumerate}
\item Revise the maximum likelihood estimates of $P$:
\begin{eqnarray*}
P_{ML(1)}(w_i), P_{ML(2)}(w_i|w_i-1), \cdots\\
\cdots, P_{ML(N)}(w_i|w_{i-N+1}^{i-1})
\end{eqnarray*}
which are calculated for each value of $n (1 \leq n \leq N)$.

\item Apply the back-off smoothing to integrate the revised ML
  estimates $P'_{ML(n)}(w_i|w_{i-n+1}^{i-1}) (1 \leq n \leq N)$.
\end{enumerate}

\renewcommand{\theenumi}{(\arabic{enumi})}

For each value of $n (1 \leq n \leq N)$, the maximum likelihood
estimates $P_{ML(n)}(w_i|w_{i-n+1}^{i-1})$ of N-gram probability $P$
obtained from the background corpus are revised to ${P'}_{ML}$ by the
following procedure.
\begin{enumerate}
\item {\em If the postfix $w_{i-k+1} \cdots w_i (1 \leq k < n)$ of the word sequence $w_{i-n+1} \cdots w_i$ is equal to the
  prefix $\hat{w}_p \cdots \hat{w}_{p+k-1}$ of one of the fixed phrases
  $\hat{w_p} \cdots \hat{w_q}$ then emphasize the $P_{ML}$ as follows:
  \begin{eqnarray*}
    \lefteqn{{P'}_{ML(n)}(\hat{w}_{p+k-1}|w_{p-n+k}^{p-1}\hat{w}_{p}^{p+k-2}) = }\\
    &&\beta_{n}(w_{p-n+k}^{p-1}\hat{w}_{p}^{p+k-2}) \cdot\\
    &&\gamma P_{ML(n)}(\hat{w}_{p+k-1}|w_{p-n+k}^{p-1}\hat{w}_{p}^{p+k-2})
  \end{eqnarray*}
  Otherwise, go to step \ref{step:2}.}

\item \label{step:2} {\em If the word sequence $w_{i-n+1} \cdots w_i$ is
  equal to the subsequence $\hat{w}_{i-n+1} \cdots \hat{w}_i$ of one
  of the fixed phrases $\hat{w}_p \cdots \hat{w}_{q}$ then emphasize
  only the longest N-gram probability $P_{ML(N)}$ as follows:
  \begin{eqnarray*}
    \lefteqn{{P'}_{ML(N)}(\hat{w}_i|\hat{w}_{i-N+1}^{i-1}) = }\\
    &&\beta_N(\hat{w}_{i-N+1}^{i-1}) \cdot \gamma P_{ML(N)}(\hat{w}_i|\hat{w}_{i-N+1}^{i-1})\\
  \end{eqnarray*}
  Otherwise, go to step \ref{step:3}.}

\item \label{step:3} {\em For all n($1 \leq n \leq N$), the revised probability is:
  \begin{eqnarray*}
    \lefteqn{{P'}_{ML(n)}(w_{i}|w_{i-n+1}^{i-1}) = }\\
    &&\beta_n(w_{i-n+1}^{i-1}) \cdot P_{ML(n)}(w_{i}|w_{i-n+1}^{i-1})\\
  \end{eqnarray*}
  }
\end{enumerate}
Here, $\gamma (\geq 1)$ is a multiplier that emphasizes the
selected N-grams, and $\beta_1(\epsilon) \cdots
\beta_N(w_{i-N+1}^{i-1})$ are normalized coefficients so that the
probabilities add up to one.

\begin{figure}[t]
  \begin{center}
    \scalebox{0.55}{\includegraphics{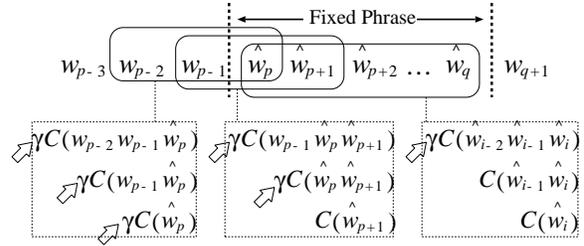}}
    \caption{Emphasizing trigram counts.}
    \label{fig:efp}
  \end{center}
\end{figure}

This can be seen as the task adaptation process by maximum a
posteriori probability (MAP) estimation \cite{federico/1996/bayesian},
in which the N-gram subset corresponding to the fixed phrases is used
as task specific data for adaptation. That is, ${P'}_{ML}$ is
equivalent to the maximum likelihood estimate calculated as follows,
\[
{P'}_{ML(n)}(w_i|w_{i-n+1}^{i-1}) = \frac{{\bf C'}_n(w_{i-n+1}^i)}{\sum_{w_i} {\bf C'}_n(w_{i-n+1}^i)}
\]
where the N-gram counts ${\bf C'}_n$ of each value of $n (1 \leq n
\leq N)$ are obtained by emphasizing the selected subset of the original
N-gram counts $C$, as shown in Fig. \ref{fig:efp}.

\section{Question Answering Engine}
\label{ss:question_answering}

\subsection{Question Answering as a Search Problem}

The question answering process is often seen as the sequence of the
question analysis, the relevant document (or passage) retrieval,
answer extraction and answer selection processes.  In this paper, we
recast these processes as a search problem.
\begin{description}
\item [Question Answering] Given query $q$ and document set
  $D$, from all the substrings in $D$, $S =
  \{(d,p_s,p_f)| d \in D, p_s < p_f; {\mbox p_s \mbox{~and~} p_f
    \mbox{~are positions in~} d}\}$, by using a evaluation function
  $L(a|q)$ defined on $a \in S$, search the most appropriate answer
  $\hat{a}$ such that $\hat{a} = \mbox{argmax}_{a \in S} L(a|q)$.
\end{description}
This defines the problem of finding a single best answer, which
corresponds to the factoid question in TREC and the subtask 1 of NTCIR
Question Answering Challenge (QAC).

\begin{figure*}[t]
\begin{center}
  \scalebox{0.9}{\includegraphics{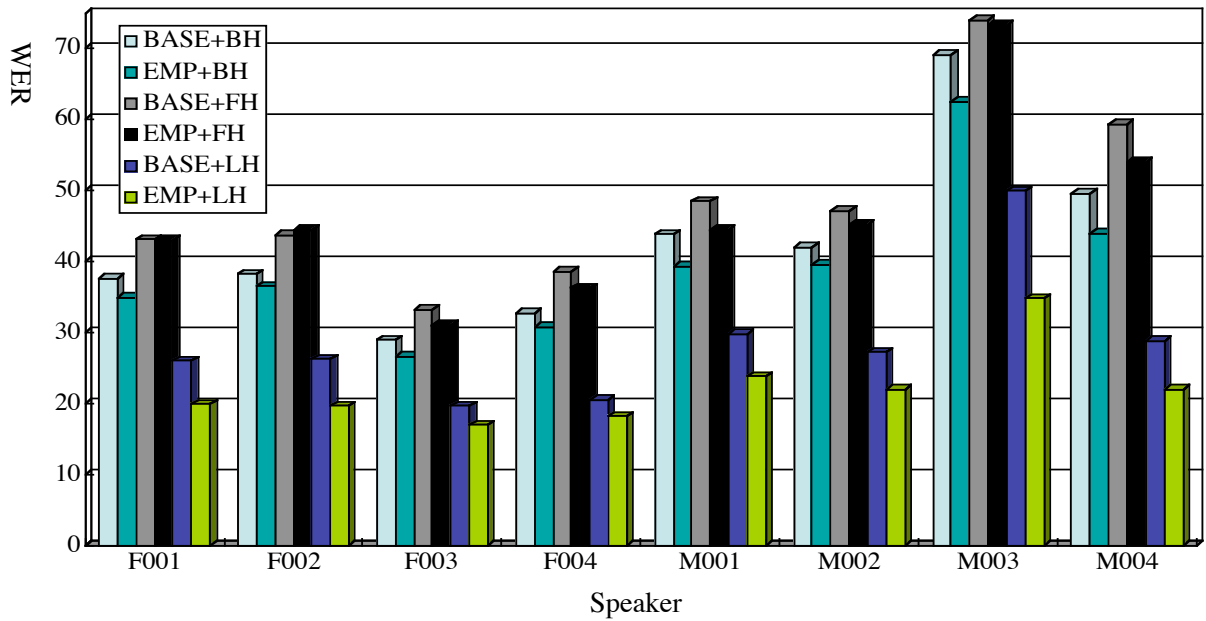}}
  \caption{WERs on spoken questions (BASE: baseline method, EMP: our method).}
  \label{fig:wer}
\end{center}
\end{figure*}

\subsection{Passage Retrieval}

The evaluation function $L$ is constructed in various aspects.  One of
them is the similarity between the question and the context of an
answer candidate. Selecting the context, or passage retrieval, is one
of the common research topics for question answering
\cite{tellex/2003/quantitative}.

Because by definition speech-driven question answering accepts a
result of speech recognition as an input, which often includes errors,
the passage retrieval must be robust against those errors.  We propose
a dynamic passage retrieval method that can accept an input including
misrecognized words.

Suppose, from given query $q$, we select the context of an answer
candidate $a$, which belongs to sentence $s_i$ of document $d = s_1
s_2 \cdots s_i \cdots s_n$. Let ${s'}_i = s_i - \{a\}$, $h$ be the
headline of $d$, and $t$ be the string ``{\it Kotoshi Kongetsu Kyou}''
(this year, this month, today). Given a number $k > 0$, let $S_i = \{
h, t, s_{i-k}, \cdots, s_{i-1}, {s'}_i, s_{i+1}, \cdots, s_{i+k} \}$.
The optimal context $\hat{C_i}$ is selected from $C_i \in 2^{S_i}$ by
maximizing the following evaluation measure $F(C_i)$.
\begin{eqnarray*}
F(C_i) & = & \frac{1 + \beta^2}{\frac{\beta^2}{R(C_i)} + \frac{1}{P(C_i)}}\\
R(C_i) & = & \frac{\mbox{Score}(q \wedge C_i)}{\mbox{Score}(q)}\\
P(C_i) & = & \frac{\mbox{Score}(q \wedge C_i)}{\mbox{Score}(C_i)}\\
\end{eqnarray*}
Here, $\mbox{Score}(A)$ is a sum of the IDFs (inverse document
frequencies) of the elements in $A$ and $\mbox{Score}(A \wedge B)$ is
a sum of the IDFs of the elements appeared commonly in $A$ and $B$.

We used $k=1$ for our experiments.  The measure $F$ corresponds to the
(weighted) F-measure often used in IR research.  The recall is more
influential than precision in calculating the F-measure, if the value
of $\beta$ is more than one.  Because the recall is important for
selecting answer candidates, we set $\beta = 2$.

\section{Evaluation}
\label{ss:evaluation}

\subsection{NTCIR Question Answering Challenge}

The test collection constructed in the first evaluation of Question
Answering Challenge (QAC-1) \cite{fukumoto/2003/question}, which was
carried out as a task of NTCIR Workshop 3, was used as the test data
for our evaluation. The task definition of QAC-1 is as follows.

Target documents are two years of Japanese newspaper articles, from
which the answers of a given question must be extracted.  The answer
is a noun or a noun phrase, e.g., person names, organization names,
names of various artifacts, money, size and date.  Three subtasks were
performed in QAC1, among which the subtask 1 is defined as follows.
\begin{quote}
  System extracts at most five answers from the documents for each
  question. The reciprocal number of the rank is the score for the
  question. For example, if the second answer candidate is correct, the score
  is 0.5.
\end{quote}
This definition is almost equivalent to the factoid question answering
in TREC. The 200 queries were used for the formal evaluation, in which
no answer was found for four questions in the target documents.  Mean
Reciprocal Rank (MRR) of the 196 queries was used to evaluate the
performance of participant systems.

\subsection{Experimental Results}

The effects of language modeling on question answering were
experimentally investigated. We extracted N-gram counts from newspaper
articles in 111 months.  The vocabulary size was 60,000.  We produced
a word network for the Japanese fixed phrases used for question
sentences. From the network, we extracted the 172 fixed phrases
accepted by the network.

We used the N-gram model produced only from the newspaper articles as
the baseline ({\em BASE}). For our proposed method, we emphasized the
N-gram counts corresponding to the fixed phrases, and produced the
adapted model ({\em EMP}). The magnification parameter $\gamma$ was
set to 50, which had been determined by our previous experiments
\cite{akiba/2003/adapting}.

All of the 200 questions in the QAC-1 test collection were used for
our experiments. We produced our spoken question data set.  The
questions were read by four females (F001, F002, F003 and F004) and
four males (M001, M002, M003 and M004). An existing LVCSR system
\cite{lee/2001/julius} was used for the purpose of transcription.

The WERs of the results of speech recognition are shown in Figure
\ref{fig:wer}. {\em BH} denotes WERs for an entire sentence, while
{\em FH} and {\em LH} denote WERs for the first and latter halves of a
sentence, respectively. We divided each sentence into the first and
latter halves by using Japanese WH-words as the boundary (the latter
half must include the WH-word), and investigated the WERs of both
halves independently. Note that the latter halves roughly correspond
to the fixed phrases used in interrogative questions. Figure
\ref{fig:wer} suggests that the proposed method ({\em EMP})
significantly decreased the WER for the fixed phrases ({\em LH}),
while it did not decrease the WER for the other parts of the input
sentences ({\em FH}).

\begin{figure}[t]
\begin{center}
  \scalebox{0.45}{\includegraphics{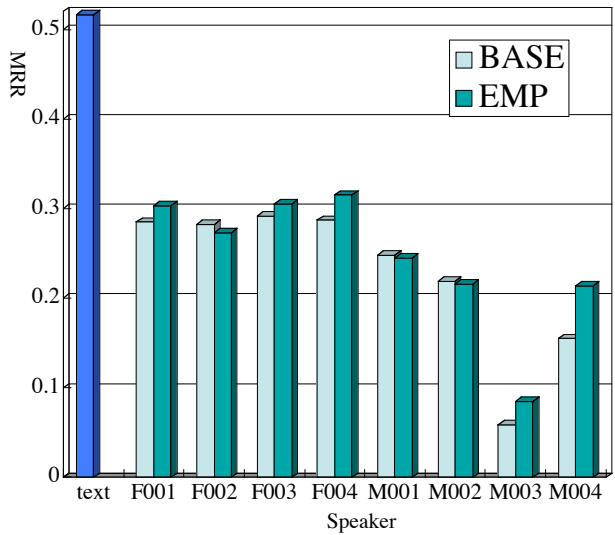}}
  \caption{QA performance by spoken questions.}
  \label{fig:qa}
\end{center}
\end{figure}

Figure \ref{fig:qa} shows the result of question answering using both
text inputs, which correspond to the speech inputs with no error, and
speech inputs by the eight speakers. For speech inputs, the best
hypothesis from the LVCSR system was used as the input of the question
answering engine. The result shows that the speech input decreased the
performance almost by half. However, when using the proposed method of
language modeling ({\em EMP}), the MRR was increased by 0.03 points on
average.

We used the paired t-test for statistical testing, which investigates
whether the difference in performance is meaningful or simply due to
chance. We found that the MRR values for BASE and EMP were
significantly different (at the 5\% level).

\section{Conclusion}

In this paper, we proposed a speech-driven question answering (QA)
system and evaluated its performance, focusing mainly on the effects
of language modeling. For evaluation purposes, we used the test
questions in the NTCIR collection, read by eight human subjects. The
experimental results showed that our language modeling method improved
the accuracy of recognizing spoken questions and consequently the
accuracy of question answering.  At the same time, when compared with
text-based QA, the performance of speech-driven QA system was not
satisfactory from a practical point of view. Future work includes
improving each module through a glass-box error analysis and extending
our system to spontaneously spoken questions
\cite{akiba/2004/collecting}.

\small
\bibliographystyle{abbrv}

\end{document}